\documentclass[]{ceurart}

\usepackage[normalem]{ulem}
\usepackage{linguex}
\usepackage{siunitx}

\begin{document}

\copyrightyear{2026}
\copyrightclause{Copyright for this paper by its authors.
  Use permitted under Creative Commons License Attribution 4.0
  International (CC BY 4.0).}

\conference{CMN’26: Workshop on Computational Models of Narrative, June 8-10, 2026, Madrid, Spain.}

\title{Comparing British and American Audio Description of Movies}

\author[1]{Igor Sterner}[%
email=igor.sterner@ed.ac.uk,
]

\address[1]{School of Informatics, University of Edinburgh, United Kingdom}

\author[1]{Alex Lascarides}[%
email=alex@inf.ed.ac.uk,
]

\author[1]{Frank Keller}[%
email=keller@inf.ed.ac.uk,
]

\begin{abstract}
Narrating the visual component of movies is known as audio description. It is a narrative technique designed to enable blind and visually impaired individuals to follow the story. However, it is far more constrained than most narratives: the descriptions not only need to convey the story in the movie, but they must also fit into gaps between dialogue and they need to conform to guidelines that exist in each region. In this work, we compare audio description created in the United Kingdom against audio description created in the United States. We use guidelines written for these two regions, alongside the impressions from a practitioner in the field, to motivate specific hypotheses about the differences. We test these hypotheses against our pre-existing corpus, which provides both human-authored American and British audio description for each of 206 movies. Results provide quantitative evidence to uphold all tested hypotheses, including differences in lexicon, the use of the progressive aspect, the use of passive constructions, the use of subjective adjectives and modifiers, when characters are named, how scenes are cued, and degree of overlap with movie dialogue and music. Our work offers a quantitative lens into the narrative technique of audio description.
\end{abstract}

\begin{keywords}
movies \sep
audio description \sep
guidelines \sep
narrative \sep
dialect
\end{keywords}

\maketitle

\section{Introduction}

Movies tell stories using a carefully curated mix of linguistic, auditory and visual information.
Audiences extract meaning and narrative from all of this information, often to great enjoyment.
However, being able to articulate that meaning-making process is a challenge.

This is a challenge tackled by accessibility services that cater to individuals for whom one or more modality is less available.
The accessibility service that aims to cater to blind and visually impaired individuals is called audio description (AD).
AD is a translation of the visual (and unclear aural) information into a verbal narration.
The narration is constrained to fit in gaps between dialogue, and it is voiced by a neutral narrator.

The task requires an audio describer to begin with a full understanding of the narrative being told \cite{vercauteren-2016-narratological}, which involves analysis of the full movie and potentially also a screenplay, dialogue subtitles, and the other available resources.
They must then identify suitable gaps between dialogue and find ways to translate the visual part of the narrative into a verbal narrative.
Every AD will be different, because these stages involve subjective interpretation and storytelling.

AD is arguably a literary and narrative art form \citep{snyder-2008-audio}.
There is debate about how much creative license should be afforded to audio describers.
This debate spans a scale of strictly objective description of the visuals \cite{adc_2009_guidelines} to a full re-narrativisation of the movie that prioritises coherent storytelling \cite{kruger-2010-audio_narration}.
For practical purposes, some parts of this debate are answered for particular regions by their AD guidelines.
For instance, American guidelines \citep[Annexe~4 of ][]{rai-2010-comparative} stipulate that all subjective interpretation is unacceptable, while British guidelines \citep[Annexe~6 of ][]{rai-2010-comparative} state that when it reduces potential confusion, particular visuals can be subject to a certain amount of interpretation.
Ultimately, however, all perception is interpretive and complete objectivity is simply impossible.
Audio describers rely on their training, experience, and common sense in order to maximise the understanding and enjoyment (i.e., access) that they provide to their audience.
Given the overarching goal of providing AD is achieving access, the extent to which American and British AD scripts differ in practice has never been quantitatively evaluated.

Consider the AD of the Trinity nuclear test scene in the 2023 movie Oppenheimer.\footnote{The video of this scene with American AD is available here: \url{https://adawardsgala.org/wp-content/uploads/2024/07/Oppenheimer.mp4}.}
{
\small
\ex. \label{ex:intro}
\a.  Kistiakowsky is open mouthed, staring through his goggles. In the car, Feynman screws up his eyes from the light. Oppenheimer takes off his goggles. An immense fireball billows into the night sky. Oppenheimer's eyelids flicker open. The explosion climbs higher and higher, white heat at its core darkening at the edges into an orange mushroom shape. Lawrence is standing outside the car with his hand on the door, watching the spreading glow on the horizon. Oppenheimer stares into the blackened orange clouds with sparks flying in their midst. On his mattress, Rabi shifts onto his side for a clearer view, keeping the tinted glass in front of his face. Other observers sit up to watch the still-blossoming explosion. Rabi starts to lower his glass. Oppenheimer watches the light fade to a red glimmer. \hfill [UK] \label{ex:intro:uk}
\b. Now, standing outside the car, Lawrence gapes as the bright line shines on his face, and Feynman squeezes his eyes shut. Teller's lips puff out with a sharp exhale as his expression remains unchanged. In the bunker, Oppenheimer removes his goggles, blinking rapidly and squinting to see the explosion growing into a column of fire that blooms high into the dark sky, taking on a mushroom shape. Seen from over Lawrence's shoulder, the explosion's brilliant light gives off an amber halo around its irregular edges, which, miles away, tinges Oppenheimer's awed face. The roiling flames billow over each other, black smoke starting to form within the inferno. Rabi finally rolls over and peers through his welder's glass at the explosion. The amber glow on Oppenheimer's face dims as the black smoke grows in the flames, which fold over on themselves. Rabi lowers the glass from his eyes as the blackness consumes the remaining flames, leaving behind a ghostly cloud of smoke in the dark.  \hfill [US] \label{ex:intro:us}

}
Both the British and American versions (\ref{ex:intro:uk} and \ref{ex:intro:us}, respectively) describe similar events in the scene, using language that is largely an objective description of the visuals.
Nevertheless, the language is almost poetic and paints a vivid scene.
There is some flexibility, especially in a short thread, as to when an action is described---e.g., compare the description of Lawrence outside a car watching the explosion (7th sentence in UK, 1st sentence in US).
There is also some flexibility on which of the less salient actions to describe, as we can see from the choices to describe only one of Kistiakowsky's or Teller's expressions.

A part of the process of creating AD also relies on knowledge of who the audience will be.
AD created in the United States will follow guidelines from that region; the same holds for AD created in other regions, in particular the United Kingdom.
In our study, we will only consider English AD from these two regions, since they form the largest available data, but English AD is also created for other regions (e.g.,~Australia) and for many other languages.
Each region will have dialectal and lexical preferences, and audio describers naturally take this into account in their decisions on how to express content.

Comparing the differences between versions of AD can offer a lens into AD as a narrative technique.
Prior work has relied on theoretical analyses of guidelines \citep{bittner-2012-guidelines} and anecdotal evidence \cite{campbell-2025-crossing}.
They offer hypotheses as to the differences between American and British AD, but do not provide quantitative evidence in support of these hypotheses.
In this paper, we aim to address this gap.
For fair and controlled experiments, we need to compare AD of the same source material from each region.
In separate work, we have collected a corpus of AD transcripts that makes this possible; see \cite{sterner-2026-reframed} for more details on the corpus and on the automatic generation of AD.
For each of 206 movies, the corpus provides a professional American AD transcript and a professional British AD transcript.

We will begin by describing hypotheses about the differences between AD in the regions. These are based on lexical variation, aspect, voice, subjectivity, character naming, dialogue overlap and description of scene changes.
The specific hypotheses we define are motivated by the AD guidelines and claims made by a leading practitioner in the creation of AD.
We design and implement automatic methods that test these hypotheses, and apply them to our corpus.
Results provide supporting evidence to uphold all of the hypotheses.
We finish by discussing the implications of our work.

\section{Hypotheses}

We make hypotheses of the differences between American and British AD based on two sources.
The first source is the AD guidelines from each region.\footnote{Rai \cite{rai-2010-comparative} appends a snapshot of guidelines at the time of their writing to their appendix (the American guidelines we use are in Annexe 4 and British guidelines in Annexe 6).}
AD guidelines provide recommendations as to how audio describers should produce their descriptions.
Our interpretation of the guidelines and the differences between them is based on Bittner's analysis \cite{bittner-2012-guidelines}.
The second source is Campbell's claims in \cite{campbell-2025-crossing}.
Campbell works in a leading role in the creation of AD at Red Bee Media, a large broadcasting and media services company.
Her claims represent the perspective of an AD practitioner.

For this section, we will provide examples from one movie, Oppenheimer (2023).
Its American AD was a finalist in both the EGA Hermes Awards and Audio Description Awards Gala.
The first unique aspect of this movie is that the officially released British AD is a post-edit of the American AD, which is uncommon (henceforth, we'll refer to this version as `post-edit').\footnote{This is a presumption, based on observing the data and both stating that they are created by the same company, Deluxe. We presume that the American AD version of Oppenheimer was provided as a starting point to a team that scripted and voiced the British AD version.}
The second unique aspect of this movie is that an entirely new UK AD version was created by the BBC for its broadcasting of the movie (henceforth, we'll refer to this version as `UK').

Consider the following scene.
Oppenheimer has been in the New Mexico desert with Lawrence and Frank.
Oppenheimer says `Come', and leads them off.
When the {\bf scene changes} to one later in the day in a tent, the AD versions are as follows.
{
\small
\ex.\label{ex:tent}
\a. Inside a tent, Lawrence drinks from a bottle and passes it to Frank. \hfill [UK]
\b. After nightfall, warm lantern light glows on the men's faces inside a tent. Lawrence passes a liquor bottle to Frank, who takes a swig. \hfill [US]
\c. \sout{After nightfall} \textbf{In a tent at night}, warm \sout{lantern} light glows on the men's faces \sout{inside a tent}. Lawrence passes a \sout{liquor} bottle \textbf{of booze} to Frank, who takes a swig. \hfill [post-edit]

}
All three versions cue the scene-change with a fronted adverbial (`inside a tent'; `after nightfall'; `in a tent at night').
Both British and American AD guidelines state that scene changes should be cued, in particular when it is difficult to infer from the soundtrack.
The American guidelines are more explicit in this recommendation, articulating that markers like `now' can be used. In Oppenheimer, `now' is used 16 times in the American version, almost all of which are removed in the post-edit (leaving 3), and also rarely used in the UK version (used 3 times).
In contrast, the British guidelines do not provide any explicit suggestions as to how signalling a change in scene should be achieved.
We can also see {\bf lexical variation} in how the alcoholic beverage is described (UK and post-edit: `a bottle' or `a bottle of booze'; US: `liquor bottle').
This difference is raised by Campbell, but not by any of the region-specific guidelines.

Consider now the following description, which occurs at the end of a scene when Oppenheimer is giving a lecture in Holland.
{
\small
\ex.\label{ex:lectern}
\a. Oppenheimer is standing at the lectern. \hfill [UK]
\b. Oppenheimer stands at a lectern. \hfill [US]
\c. \textbf{At the front,} Oppenheimer stands at a lectern \textbf{looking slightly nervous}. \hfill [post-edit]

}
The British and American descriptions are near-identical, except for the fact that they differ in {\bf aspect}.
Both are in the present tense, but the British description employs the progressive aspect, while the American description is simple.
Campbell claims that American descriptions employ almost exclusively the simple present tense, while in British descriptions there is the option to switch to the present progressive and simple present for variation.
This claim is supported by the guidelines, which all state that the present tense should be used, with the guidelines from the UK and from Ireland additionally allowing progressive aspect to be used.

In the post-edited version of the American description, the describer has also chosen to incorporate a more detailed description of Oppenheimer's body language (`looking slightly nervous').
Note that the describer has chosen not to describe the visuals directly (e.g. Oppenheimer's pursed lips or avoidant eyes).
Instead the adjective `nervous', additionally modified by the adverb `slightly', is used to make a judgment about Oppenheimer's emotions.
In this case, this decision was presumably made due to the very short gap between dialogue into which this description has to fit.
The British guidelines explicitly afford flexibility with regard to {\bf subjective interpretation}, in particular when this reduces potential confusion.
However, the American guidelines are explicit in discouraging this.
Campbell acknowledges the assumption that British AD tends to do more work in interpreting emotions than American AD.
However, she argues that it is impossible to be entirely objective and there is not a claim of any systematic difference.

Consider now a scene when Oppenheimer is walking down a dark street to an event and an FBI agent is taking notes of the license-plates of attendees.
{
\small
\ex.\label{ex:torch}
\a. Oppenheimer sees a torch being shone at cars parked on a suburban street.\hfill [UK]
\b. Oppenheimer notices someone with a flashlight checking out a car and taking a note. \hfill [US]
\c. Oppenheimer notices someone \sout{with} \textbf{shining} a \sout{flashlight, checking out} \textbf{torch by} a car and \sout{taking a note.} \textbf{making notes.} \hfill [post-edit]

}
The British description employs the {\bf passive voice}, while the American description stays in the {\bf active voice}.
Campbell argues that in British AD the passive voice is used as a tool to direct the audience's focus, while American AD typically avoids it.
The guidelines do not suggest any major difference in syntactic construction between American and British AD.

Consider now how AD {\bf names characters}.
In Oppenheimer, there are three board members in the hearing for his security clearance.
We see the third board member, Ward Evans, at the back of the hearing from the beginning of the movie, but he is not named in the dialogue until towards the end of the movie (one of the other board members says `A full written opinion, with a dissent from Mr. Evans, will be issued to the AEC in the coming days.').
A full 8 minutes before this utterance occurs in the movie, the British description opts to name him (`A smile from Evans, the oldest board member.').
In contrast, at this point in the movie, the American description does not name him (`An elderly member of the board smiles.'). In contrast again, in the post-edit, the name is introduced (`An older member of the board, Ward Evans, smiles.').
The American version names him for the first time directly after the dialogue that introduces him (`Robb shakes hands with the chairman, the elderly Evans and the third board member.').
The guidelines differ in whether they permit characters to be named before the sighted audience finds out their name: while the American guidelines stipulate that the description should always wait to name the character after the sighted audience finds out their name, British guidelines state that it is also acceptable to name a character earlier, unless there is mystery attached to the name.
Campbell argues that human describers use their own common sense to prioritise proportionate access to that of a sighted person.

The creation of an AD script also involves selecting {\bf suitable moments to voice the description}.
The British guidelines stipulate a strict rule to never voice description over dialogue.
Meanwhile, the American guidelines are somewhat more permissive.
As one example, when Oppenheimer is left behind in a lab and his classmates leave, one of them shouts back “Don't forget to clean up!”.
The American version voices description over this entirely, while the British version does not.
The American description opts to talk about a green apple which soon becomes salient to the narrative.

In sum, claims of the differences between American and British audio description are as follows.

\begin{enumerate}
    \item \textbf{American terms}: American AD uses a higher frequency of American English lexical items.
    \item \textbf{British terms}: British AD uses a higher frequency of British English lexical items.
    \item \textbf{Present progressive}: British AD uses the present progressive aspect more frequently.
    \item \textbf{Subjectivity}: British AD exhibits greater subjectivity in interpreting visual information and character emotions.
    \item \textbf{Passive}: British AD uses the passive voice more frequently.
    \item \textbf{Naming in advance}: British AD names characters in advance more frequently.
    \item \textbf{Dialogue overlap}: American AD voices over dialogue (and music) more frequently.
\end{enumerate}

\section{Methods}

\subsection{Materials}

We now test these claims using our pre-existing corpus of AD \cite{sterner-2026-reframed}.
For each of 206 movies, the corpus provides professional American and British AD, transcribed from their audio tracks with automatic methods.\footnote{There is an automatic step that supports the correct spelling of character names. Example~\ref{ex:intro} is derived from the same automatic transcription, which resulted in the correct spellings of the characters.}
Automatic signal processing steps were taken to ensure that both tracks describe identical versions of the movie and to align them with each other.
For every movie, the corpus also provides human-transcribed and automatically aligned dialogue subtitles.

There also exists an accessibility service designed to cater to individuals who are deaf and hard of hearing, which is often called SDH (Subtitles for the Deaf and Hard-of-Hearing).
These subtitles contain salient non-verbal audio (e.g.,~Lawrence exhaling during the Trinity explosion scene), unclear speaker identification (e.g.,~that Oppenheimer is narrating immediately after), and text simplification when dialogue is spoken too quickly to read (from a different movie, `I just thought that maybe this was gonna be more effective.' becomes `I just thought this would be more effective.').
All 206 movies are also provided with these subtitles.
The corpus additionally provides a number of video excerpts released publicly for the movies, as well as aligned screenplays.

All automatic tools used in the construction of this dataset were validated against a separate set of ten full movies, for which we gathered professional transcription, manually annotated scene boundaries and manually annotated screenplay alignment.
Scene boundary annotation was performed first based on the movie visuals, and then the British and American AD corresponding to each scene was determined.
For our experiments, we will only use the corpus American AD, British AD, and the dialogue subtitles.\footnote{The exception to this will be our analysis of scene changes, for which we will only use the ten manually processed movies.}
More details about the corpus and its construction are in \cite{sterner-2026-reframed}.

\subsection{Automatic Pre-processing}
We process each AD and dialogue sentence with automatic NLP tools from the Stanza toolkit \cite{qi-2020-stanza}.
This provides syntactic parses in which the output features Universal Dependencies \cite{marneffe-2021-ud}, Penn Treebank POS tags \cite{marcus-1993-building} and named entity recognition according to \cite{weischedel-2013-ontonotes}.
Details on how we use the parsed output of the different versions of AD to estimate the frequency of the linguistic features of note are given below.

\subsection{Significance Testing}

We test differences between American and British AD for statistical significance with paired, two-tailed permutation tests, approximated by $N=10{,}000$ and with threshold $\alpha=0.05$.
The intuition for this statistical test is that if there was not a systematic difference between American and British AD, we could randomly swap the `American' and `British' labels for any given movie AD without changing the average results.

\subsection{Statistics}

Table~\ref{tab:corpus-statistics} gives some statistics of the AD and subtitle portions of the corpus.
Some differences between the British and American versions are already evident.
On average, the AD of a single movie contains 588 and 831 sentences in the two versions, respectively.
We observe significantly higher lexical diversity\footnote{We used the length-controlled variant of type-token ratio from McKee \cite{mckee-2000-measuring}.} and mean dependency distance in British AD, as compared to American AD.
Lexical diversity is significantly lower than in the subtitles.
The mean speech rate in both versions is higher than the 160 words per minute suggested in the American guidelines.
Finally, we observe that in AD, characters are mentioned significantly more than in dialogue ($p<0.01$ for AD variants).
This aligns with the statement in the British guidelines that names tend to be used more frequently in AD than in normal speech.

\begin{table}
\centering
\caption{Corpus statistics. The subtitles are not provided with sentence segmentation, nor do we rely on such segmentation for any analysis on the subtitles in this paper. $^*$This number is artificially c. 4\% lower for some of the movies, due to a step in our alignment that corrects the different speed at which movies are shown in the two regions.}
\label{tab:corpus-statistics}
\begin{small}
\begin{tabular}{l S[table-format=4.2] @{\hspace{1.5cm}} S[table-format=4.2] @{\hspace{1.5cm}} S[table-format=4.2]}
\hline
& {\textbf{UK}} & {\textbf{US}} & {\textbf{Subtitles\qquad \  \ }} \\
\hline
\# movies & 206 & 206 & 206 \\
\# tokens  & 1137598 & 1419576 & 1899006 \\
\# sentences & 121123 & 171224 \\
\# tokens per movie & 5522$_{\pm2064}$ & 6891$_{\pm2323}$ & 9218$_{\pm3397}$  \\
\# sentences per movie  & 588$_{\pm217}$ & 831$_{\pm237}$ \\
Lexical diversity & 87.2$_{\pm10.1}$ & 84.5$_{\pm9.6}$ & 113.9$_{\pm12.5}$\\
Mean dependency distance & 2.5$_{\pm0.1}$ & 2.4$_{\pm0.1}$\\
Speech rate (tokens per minute) & 190.7$_{\pm13.1}$$^*$ & 194.2$_{\pm12.5}$ & 167.0$_{\pm21.13}$ \\
Number of character names per movie & 340.6$_{\pm136.1}$ & 448.8$_{\pm161.5}$ & 217.0$_{\pm101.7}$ \\
\hline
\end{tabular}
\end{small}
\end{table}

\subsection{Experimental Tests}

\paragraph{Lexical variation.}

We quantify lexical difference by measuring the frequency of American- and British-specific terminology.
Lists of British and American terms are taken from online resources.\footnote{American terms: \url{https://en.wikipedia.org/wiki/Glossary_of_American_terms_not_widely_used_in_the_United_Kingdom}, British terms: \url{https://en.wikipedia.org/wiki/Glossary_of_British_terms_not_widely_used_in_the_United_States}.}
There are 314 American terms and 677 British terms.
Some of these terms are only more frequent in one region for a particular sense, and our subsequent analysis will be sense-independent.
As one step to address this, we remove from the American list words also in the top 5,000 most frequent words in the British National Corpus \cite{leech-2001-bnc_frequencies}.
Similarly, we remove from the British list the top 5,000 most frequent words in the Corpus of Contemporary American English \cite{davies-2008-coca}.
This results in removing 3 and 14 words, respectively, leading to lists of length 311 and 663 words.
Furthermore, for each movie we exclude all terms matching or contained in character names in the movies (for instance in one movie, a character is called `ladybug', which is removed from the American list for that movie).
We compute the frequency of these terms per sentence within the AD transcripts and report a macro-average over movies.

\paragraph{Aspect and voice.}

For differences in aspect, we identify instances of the present progressive by detecting when all of the following conditions are met in a sentence: (1) a gerund or present participle (2) has an auxiliary dependent form of `to be' (3) which is in the present tense.
We compute the proportion of such description sentences in each movie and report the macro-average.
We identify sentences that use the passive voice by labelling sentences that contain any dependency relation marking a passive subject or passive auxiliary.
We compute the proportion of description sentences containing passive constructions in each movie and report the macro-average.

\paragraph{Subjective judgments.}

For our automatic analysis of subjective judgments in AD, we focus on the use of adjectives and how they are modified.
De Smedt \cite{de-smedt-2012-vreselijk} annotates English adjectives with percentage subjectivity scores and provides an algorithm for incorporating scores from adverbial modifiers.
We apply their approach and determine the subjectivity of each description sentence and report a macro-average over movies.

\paragraph{Naming in advance.}

We identify whether AD introduces a character's name before they are named in the dialogue.
We use character lists from IMDb (modulo uncredited characters and enumerated characters) and define a mapping based on all unique permuted variants of the name.
The first named entity that maps to each character is then identified in each AD script and the dialogue subtitles.
We keep occurrences that match both of the following conditions: (1) the character is identified as occurring in both AD versions earlier than 60 seconds after its occurrence in the dialogue subtitles, (2) at least one of the AD versions names the character between the time the character is named in the dialogue subtitles and 60 seconds later.
After these conditions, there are three cases: (1) neither AD version names in advance, (2) the American version names in advance, (3) the British version names in advance.
We compute counts of cases (2)/(1)+(2)+(3) and (3)/(1)+(2)+(3) and report a macro average over the movies with at least one detected character matching the conditions (183/206 movies).

\paragraph{Dialogue overlap and scene changes.}

Our measure of dialogue overlap is the proportion of duration of AD that overlaps with the dialogue subtitles.
Dialogue subtitles also include music lyrics.
For our experiments on how scene changes are articulated, we only use the separate set of ten movies with manual scene annotation.
Based on this annotation, we train a BERT language model \cite{devlin-2019-bert} to identify scene boundaries based on the AD text alone.
The model takes as input six sentences, and a separator token is placed between sentences three and four.
The training task is binary classification, with labels of whether the separator token represents a scene boundary or not.
We perform cross-validation on the ten movies, and report results based on training on the American, British or both AD versions.
For fair comparisons, we need to control the amount of data available to each trained model.
Earlier we found that American AD is typically longer than British AD; training on both AD versions will always result in more data than training on only one.
We address this by subsampling the data from each movie such that the number of positive and negative samples is equal to the number from the movie with the fewest samples.
This equates to sampling 74 descriptions that start new scenes and 474 that do not.

\section{Results and Discussion}

\begin{table}
\centering
\caption{Results based on a macro average over 206 movies ($^*$only 183 movies; for the remainder, zero valid characters were detected for the analysis). In each row, the underlined result indicates the larger number of the two.}
\label{tab:main-results}
\begin{small}
\begin{tabular}{lllrrc}
\hline
& \textbf{Comparison} & \textbf{Unit} & \textbf{UK} & \textbf{US} & \textbf{Significance} \\
\hline
(1) & American terms & (terms per 1,000 sentences) & 5.0 & \underline{16.8} & $p<0.01$ \\
(2) & British terms & (terms per 1,000 sentences) & \underline{15.8} & 4.6 & $p<0.01$ \\
(3) & Present progressive & (proportion of sentences, \%) & \underline{3.0} & 0.3 & $p<0.01$ \\
(4) & Subjectivity & (mean sentence score, \%) & \underline{16.2} & 13.3 & $p<0.01$ \\
(5) & Passive & (proportion of sentences, \%) & \underline{4.6} & 1.3 & $p<0.01$ \\
(6) & Naming in advance$^*$ & (proportion of characters, \%) & \underline{31.4} & 9.0 & $p<0.01$ \\
(7) & Dialogue overlap & (proportion of AD time, \%) & 8.5 & \underline{14.1} & $p<0.01$ \\
\hline
\end{tabular}
\end{small}
\end{table}

Results of the application of the tests to the corpus are given in Table~\ref{tab:main-results}.

\paragraph{Lexical variation.}

Rows~1 and~2 provide supporting evidence of a difference in lexicon.
There are an average of 16.8 American terms per 1,000 sentences of American description, significantly more than the 5.0 in British description sentences.
The British terms are used on average 15.8 times per 1,000 sentences of British description, significantly more than the 4.6 times in American description sentences.
These frequencies are still low, in part because our lists are not comprehensive and in part also demonstrating that lexical variation only plays a small role in differences between the regions (around 1\% of sentences).

The most frequent American terms in American AD are `SUV' (414 US counts vs.\ 134 UK counts; British equivalent might be `4x4'), `flashlight' (337 US counts vs.\ 40 UK counts; `torch'), and `sidewalk' (307 US counts vs.\ 19 UK counts; `pavement').
The most frequent British terms in British AD are `amongst' (184 UK counts vs.\ 16 US; American equivalent might be `among'), `car park' (152 UK counts vs.\ 2 US; `parking lot'), and `windscreen' (138 UK counts vs.\ 6 US; `windshield').
These are unsurprising results, and a clear demonstration of why AD needs to be localised for these two dialects of English.

\paragraph{Aspect and voice.}
Row 3 provides clear evidence that the present progressive is rarely used in American description (0.3\% of sentences) but used significantly more frequently in UK descriptions (3.0\% of sentences).
Passive constructions are used more than three times as frequently in British AD compared to American AD (row 5; 4.6\% vs. 1.3\%, $p<0.01$).
Evidently these are both only tools in the toolkit of a British audio describer, albeit even then not frequently used tools.
Both the progressive aspect and the passive voice are often considered to be more complex syntax than simple aspect and active voice; algorithms that are designed to simplify English often change passive sentences to active ones.
Meanwhile the American guidelines are more explicit than the British ones in their recommendation to use simple language --- our results here can be seen as a consequence of this.

\paragraph{Subjective judgments.}

Based on this analysis, we find a significantly higher average subjectivity score in British description sentences compared to American sentences (row 4; 16.2\% vs. 13.3\%, $p<0.01$).
For instance, the word `beautiful' (which has a score of 100\% subjective) occurs almost three times more frequently in British description than in American description (87 vs. 35 occurrences).
Indeed the American guidelines directly address the use of this word, negatively stating that it is vague and evaluative.

Consider the following two uses of `beautiful':
{
\small
\ex. \label{ex:beautiful-1}
% movie: tt13375076
\a. Glancing around the room, he turns back to find a beautiful woman glaring at him. \hfill [UK]
\b. He looks away, then looks back and finds a dark-haired young woman standing before him. \hfill [US]

\ex. \label{ex:beautiful-2}
% movie: tt1473832
\a. Bridget notices the church is mainly filled with beautiful young women. \hfill [UK]
\b. Bridget notices many of the pews filled with fashionable young women. \hfill [US]

}
In example~\ref{ex:beautiful-1}, the British description opts for a more detailed description of how the character scans the room (`Glancing around the room' vs. `He looks away'), and recovers from this use of narration time by describing the woman as `beautiful' (vs. `dark-haired' and `young').
In example~\ref{ex:beautiful-2} we can see that the two versions are near-identical, except for the choice to use `beautiful' (British description) as compared with the one syllable longer `fashionable' (American description).

\paragraph{Naming in advance.}

Row 6 gives evidence that in British AD characters are consistently named in advance more frequently (31.4\% vs. 9.0\% of characters identified in our analysis, $p<0.01$).
When a character is named in advance in British AD, they are named on average 11 minutes early, compared to seven minutes in American AD.
At the extremes, on one occasion a character is named in British AD over 100 minutes before being named in the dialogue, and on another occasion a character is named over 60 minutes early in American AD.
Both cases concern supporting characters that appear only briefly in the opening quarter of the movie but are more fully developed in the final quarter of the movie.
Our findings show that some American audio describers do opt to name in advance, even if this goes against their guidelines.
Note that the cases under consideration here only concern when one, and not both, of the versions opted to name in advance.
The 9.0\% of characters named in advance in American AD thus represents occasions when the British describer did not name in advance.
We suspect that this number represents roughly how often describers from the same region will differ in their choices to name in advance.
This leads to the suggestion that in 22.4\% of cases the British describer made a regionality-influenced decision to name in advance.
In most cases (59.6\%), both versions are naming the character immediately after they are named in the dialogue.

\paragraph{Dialogue overlap.}

Row 7 shows that 14.1\% of American description time overlaps with the subtitles, compared to the significantly lower 8.5\% of British description time.
This supports the claim that British AD is stricter about not overlapping with character dialogue and music.
This result is, however, unsurprising given our earlier statistic that British AD uses on average 29.2\% fewer sentences.

\paragraph{Scene changes.}

\begin{table}
\centering
\caption{Cross-validation results based on training a model to predict movie scene boundaries from AD transcripts.}
\label{tab:scenes}
\begin{small}
\begin{tabular}{lcccccc}
\toprule
& \multicolumn{3}{c}{\textbf{UK}} & \multicolumn{3}{c}{\textbf{US}} \\
\cmidrule(lr){2-4} \cmidrule(lr){5-7}
& \textbf{$P$} & \textbf{$R$} & \textbf{$F_1$} & \textbf{$P$} & \textbf{$R$} & \textbf{$F_1$} \\
\midrule
Trained on UK    & 62.6 & 55.4 & 58.4 & 61.9 & 72.4 & 65.8 \\
Trained on US    & 73.6 & 42.9 & 53.7 & 70.6 & 66.0 & 67.0 \\
Trained on US+UK & 66.1 & 55.1 & 59.6 & 65.1 & 69.0 & 65.9 \\
\bottomrule
\end{tabular}
\end{small}
\end{table}

Results are shown in Table~\ref{tab:scenes}.
We can see that recall is significantly lower for the task of identifying scene boundaries from British AD transcripts, as compared to from American AD transcripts (trained on UK: 55.4\% vs. 72.4\%, $p=0.003$; trained on US: 42.9\% vs. 66.0\%, $p=0.003$; trained on UK+US: 55.1\% vs. 69.0\%, $p=0.013$).
Training only on American transcripts and testing on British transcripts leads to the worst performance ($F_1=53.7\%$, differences to other $F_1$ scores all have $p\leq0.05$).
This suggests that this model learnt to rely on cues (most likely explicit cues like `now') in American transcripts that are not present as frequently in the British transcripts.
In sum, these results provide supporting evidence that scene boundaries are verbalised less explicitly in British AD than in American AD.

\section{Conclusion}

AD is a narrative technique.
We have shown at scale the existence of small but statistically distinguishable differences between how the narrative is verbalised in American vs.\ British descriptions.
All of the differences are justified based on differences between the regions' guidelines, except for lexical variation, which has not received prior theoretical treatment.
\begin{enumerate}
  \item American AD uses American terms roughly three times more frequently than British AD does; the same holds vice versa.
  \item American AD rarely uses the progressive aspect or passive voice, while British AD uses these in 3\% and 4\% of sentences, respectively.
  \item Both American and British AD use subjective adjectives and modifiers, but American AD uses them more sparsely.
  \item Of characters named in at least one version immediately after being named in the dialogue, American AD names that character in advance at one third of the frequency in British AD.
  \item Both American and British AD, on occasion, narrate over dialogue, but American AD overlaps more frequently (and is also longer in general).
\end{enumerate}
Our work contributes to a better understanding of the creation of AD.
It can serve to help practitioners better analyse their own habits and encourage the field to make more active decisions in guidelines.
AD also serves as an attractive domain for research in narratology, because it is a highly constrained form of storytelling that makes the underlying processes of narrative construction in movies explicit.
Using our corpus of British and American AD, we have here provided empirical evidence that AD is constrained both by dialect differences and by guidelines.

\section*{Acknowledgements}

IS is funded by an EPSRC PhD studentship (project reference 2923920).
We thank Nina Gregorio and Marie Campbell for comments.

\bibliography{myrefs}

\end{document}